\documentclass[10pt,twocolumn,letterpaper]{article}

\usepackage[pagenumbers]{iccv} % To force page numbers, e.g. for an arXiv version

\usepackage{amsmath,amsfonts,bm,xspace}

\def\eqref#1{equation~\ref{#1}}

\def\1{\bm{1}}

\def\vd{{\bm{d}}}

\def\vo{{\bm{o}}}

\def\vt{{\bm{t}}}
\def\vu{{\bm{u}}}

\def\vx{{\bm{x}}}

\def\mF{{\bm{F}}}

\def\mH{{\bm{H}}}
\def\mI{{\bm{I}}}

\def\mK{{\bm{K}}}

\def\mM{{\bm{M}}}

\def\mR{{\bm{R}}}

\DeclareMathAlphabet{\mathsfit}{\encodingdefault}{\sfdefault}{m}{sl}
\SetMathAlphabet{\mathsfit}{bold}{\encodingdefault}{\sfdefault}{bx}{n}

\def\gE{{\mathcal{E}}}
\def\gF{{\mathcal{F}}}

\def\gL{{\mathcal{L}}}
\def\gM{{\mathcal{M}}}
\def\gN{{\mathcal{N}}}
\def\gO{{\mathcal{O}}}

\DeclareMathOperator*{\argmin}{arg\,min}

\definecolor{iccvblue}{rgb}{0.21,0.49,0.74}
\usepackage[pagebackref,breaklinks,colorlinks,allcolors=iccvblue]{hyperref}

\def\confName{ICCV}
\def\confYear{2025}

\title{RayPose: Ray Bundling Diffusion for Template Views \\in Unseen 6D Object Pose Estimation} 

\author{
Junwen Huang$^{1,2}$ \and 
Shishir Reddy Vutukur$^{1,2}$ \and 
Peter KT Yu$^{3}$ \and
Nassir Navab$^{1,2}$ \and 
Slobodan Ilic$^{1}$ \and 
Benjamin Busam$^{1,2}$ \and
\\
$^{1}$Technical University of Munich ~\
$^{2}$Munich Center for Machine Learning ~\
$^{3}$XYZ Robotics \\
}
\usepackage{comment}
\usepackage{booktabs}
\usepackage{amssymb} % For symbols
\usepackage{pifont}  % For checkmark and crossmark
\newcommand{\cross}{\ding{55}}

\begin{document}
\maketitle
\begin{abstract}
Typical template-based object pose pipelines estimate the pose by retrieving the closest matching template and aligning it with the observed image. However, failure to retrieve the correct template often leads to inaccurate pose predictions. To address this, we reformulate template-based object pose estimation as a ray alignment problem,  where the viewing directions from multiple posed template images are learned to align with a non-posed query image. Inspired by recent progress in diffusion-based camera pose estimation, we embed this formulation into a diffusion transformer architecture that aligns a query image with a set of posed templates. We reparameterize object rotation using object-centered camera rays and model object translation by extending scale-invariant translation estimation to dense translation offsets. Our model leverages geometric priors from the templates to guide accurate query pose inference. A coarse-to-fine training strategy based on narrowed template sampling improves performance without modifying the network architecture. Extensive experiments across multiple benchmark datasets show competitive results of our method compared to state-of-the-art approaches in unseen object pose estimation.

\end{abstract}
\let\thefootnote\relax
\footnotetext{\hspace{-0.6cm} Project page: \url{https://demianhj.github.io/projects/RayPose}\\
\tt \{frstname.lastname\}@tum.de
}
    
\section{Introduction}
\label{sec:intro}
\begin{figure}[t]
    \centering
    \includegraphics[width=1\columnwidth]{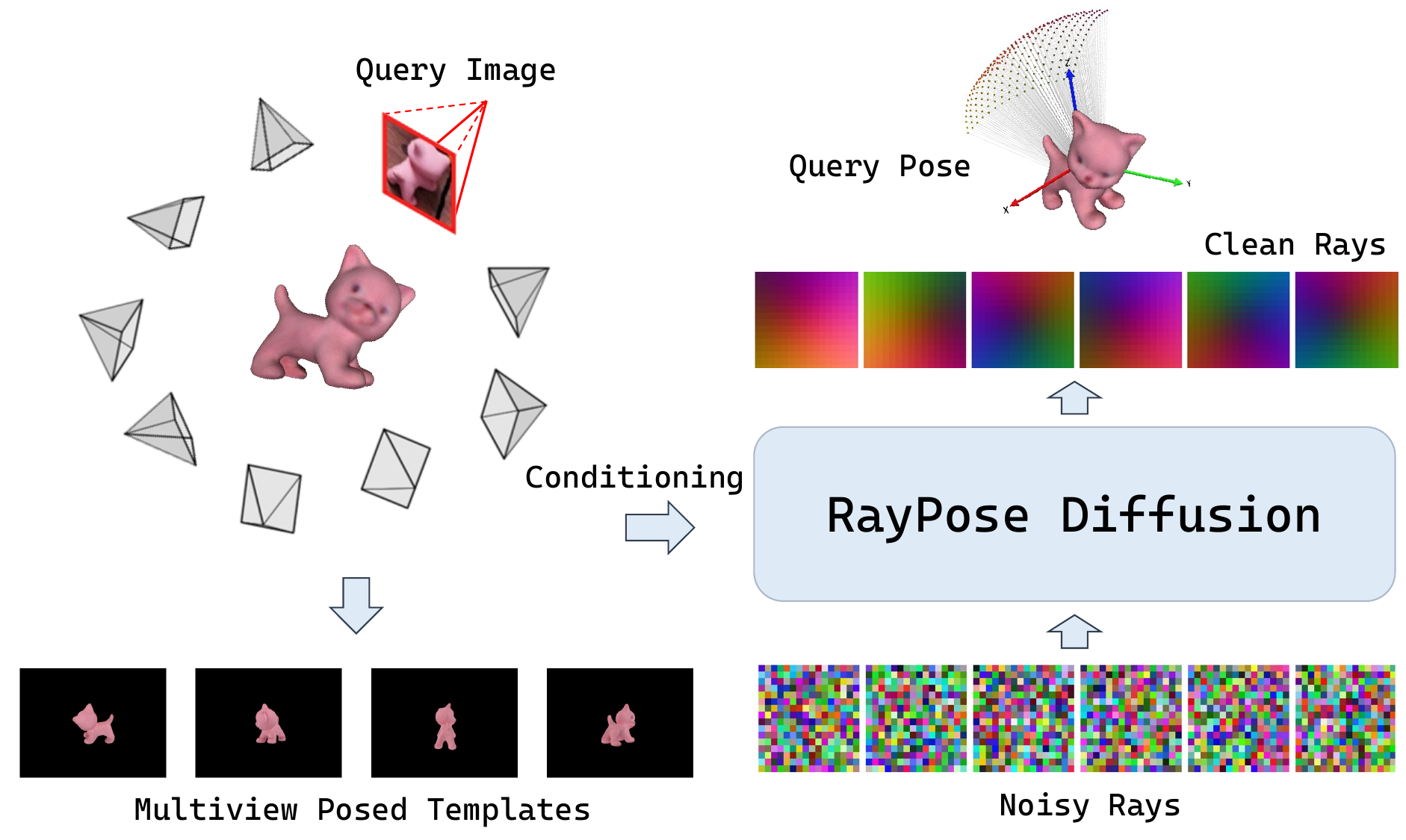}
    \caption{Given a novel object query image, our method accurately predicts the object's 6D pose using a multiview diffusion model conditioned on a set of template images with known poses. Leveraging our proposed structured 2D pose maps, represented as bundles of rays, the diffusion model recovers the query object's pose by progressively denoising these ray bundles.
}
    \label{fig:overview}
    \vspace{-0.5cm}
\end{figure}

Multi-view vision is a core element for 3D perception~\cite{hartley2003multiple}.
Spatial understanding and measurements often depends on multiple cameras or temporally-varied perspectives over time to reason about the surrounding in 3D.
Also for the task of object pose estimation -- the prediction of rotation and translation of objects in space,  multi-view constraints can be beneficial~\cite{labbe2020cosypose}.
In many computer vision applications, like robotic bin picking, augmented reality, and autonomous driving, multiple cameras or acquisitions are not available and the system needs to function even with a single monocular RGB image.

In object pose estimation literature, much effort has been put into learning other constraints, such as object appearance from visual data during training. Instance-based approaches~\cite{wang2021gdrnet,su2022zebrapose} therefore get their constraint from access to model appearance during training while category-level approaches~\cite{Wang_2019_CVPR,jung2022housecat6d,chen2024secondpose,li2025gce} use object shape and semantic priors.
Despite the excellent results that benefit from deep learning, these approaches require training for every new object or object category from scratch and creating synthetic training data from a CAD model is also computationally expensive. To overcome per-object training, researchers have been working on unseen object pose estimation with access to textured CAD models during inference ~\cite{megapose,nguyen2024gigapose,zeropose,gcpose,matchu,foundationpose,ornek2024foundpose}. These advancements promise to overcome the scalability and flexibility hurdles of object-specific approaches.

These approaches are unable to access multiple views by input design and template approaches typically solve a classification task first: which is the best template given an image query? Consecutive steps after template matching involve correspondence estimation, pose prediction, and optionally refinement~\cite{osop,nguyen2024gigapose,foundationpose}. 
Instead of finding the best possible posed template and then building pairwise correspondences, we think of the problem as an implicit bundle agreement among multiple views, using multiple template-query tuples to reason about 3D, with the advantage of having the template already posed. 

Learning to reason about 3D from multiview inputs has been extensively studied in prior work~\cite{colmap,labbe2020cosypose,lin2024relposepp,he2022oneposeplusplus}. More recently, diffusion models have emerged as powerful tools for 3D reasoning, demonstrating remarkable generalization capabilities~\cite{MVDiff,MVDiffusion,MVDiffusion+,ZeroshotNovelView,MVDream,lu2025matrix3d,wang2023pd,zhang2024raydiffusion}. Among them, PoseDiffusion~\cite{wang2023pd} addresses the inverse problem of structure-from-motion by directly diffusing camera poses within a probabilistic diffusion framework, modeling the conditional distribution of poses given input images. Building upon this, recent work~\cite{zhang2024raydiffusion} introduces an overparameterization of camera poses using Plücker coordinates~\cite{plucker1828analytisch}, representing a pose as 2D maps of ray direction and ray moment. This formulation is shown to be more compatible with diffusion processes and leads to improved accuracy in relative pose estimation. These approaches exhibit strong generalization and can infer relative camera poses even in novel scenes composed of entirely unseen images. Motivated by this capability, we propose to leverage a set of posed template images and a single query image to estimate the 6D pose of an object in the query by building on the strengths of multiview diffusion-based backbones.

Although diffusion models have shown success in relative camera pose estimation~\cite{zhang2024raydiffusion,wang2023pd}, they are suboptimal for object pose estimation due to scale differences: camera poses are defined in a large world coordinate system, while object poses reside in a compact, object-centric space. To bridge this gap, we propose novel object-centric pose representations tailored for 6D object pose estimation. For rotation, we replace camera-centric Plücker coordinates with an object-centered formulation where rays are structured as a 2D image-aligned grid. For translation, we extend the Scale-Invariant Translation Estimation (SITE) framework~\cite{li2019cdpn} to generate a dense translation map. This object-centric parameterization enables more precise and disentangled reasoning about object-level 6D pose within the diffusion framework.
Our structured pose diffusion framework takes a query image of an unseen object cropped from the scene and a set of posed images as templates, obtained by synthetic rendering from a CAD model, and generates precise 6D object pose predictions. We also propose a coarse-to-fine object pose estimation strategy by sampling the template with a narrower distribution based on the inputs. We evaluated our method on standard benchmark datasets from the pose estimation benchmark~\cite{sundermeyer2023bop} and compared it to recent methods for unseen object pose estimation. The performance of our method surpassed the results of the related works, and a detailed ablation study verified our design choices. This paper makes the following contributions:
\begin{itemize}
    \item we formulate unseen object pose estimation as ray bundling problem between multiview templates and RGB query, which helps the network to capture the correlation between query and templates in 3D space.
    \item we introduce object-centric orientation and translation over-parameterization suitable for learning within diffusion framework. 
    \item we propose a flexible diffusion-based 6D object pose framework for unseen object pose estimation that can be extended to a coarse-to-fine prediction by using different template sampling  
\end{itemize}

\section{Related Work}
\label{related}
\begin{figure*}[t]
    \centering
\includegraphics[width=0.92\linewidth]{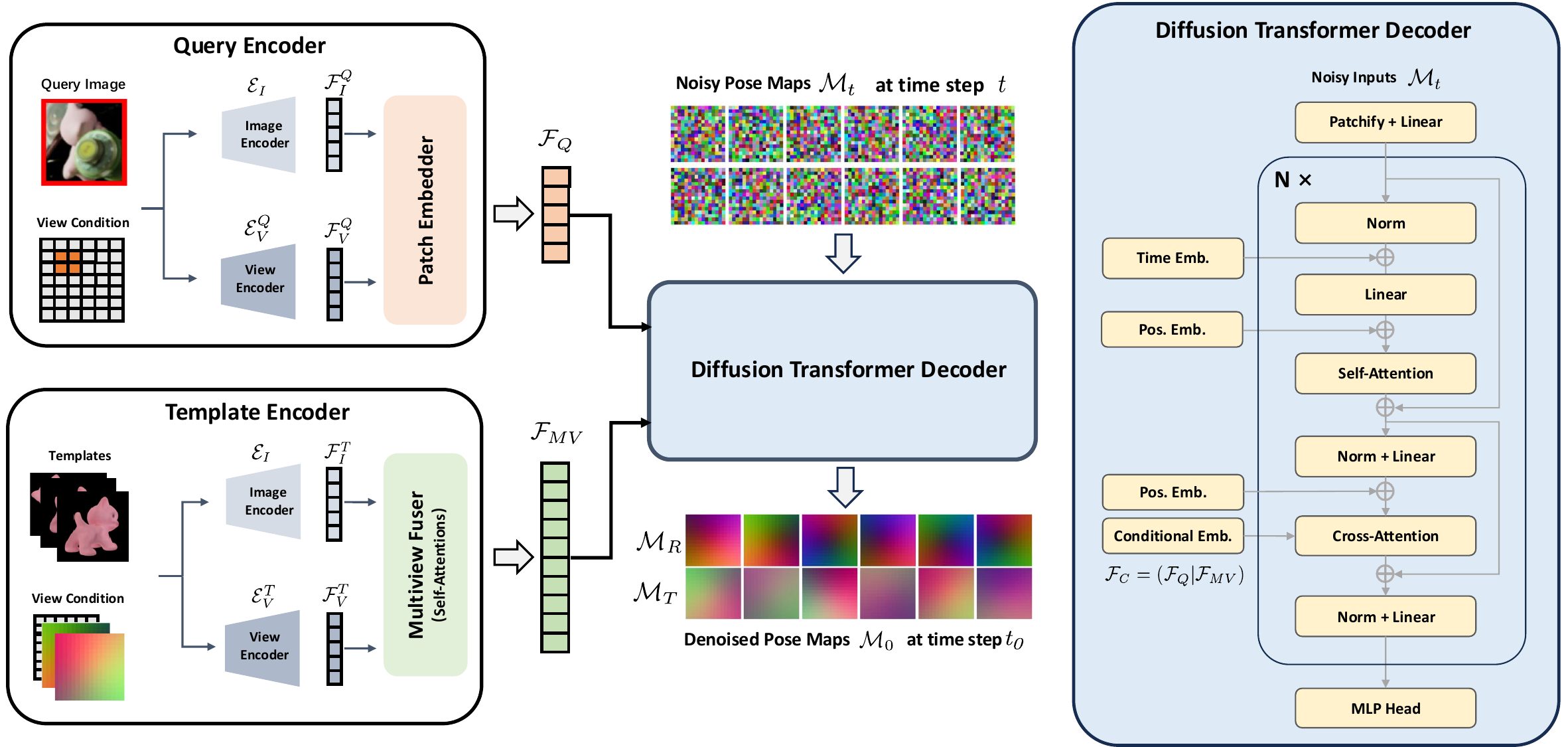}
    \vspace{-0.1cm}
    \caption{Pipeline overview of our method. We represent the 6D object pose using structured rotation and translation maps and employ a diffusion model to estimate the pose from random inputs. Given a query image of an unseen object and multiple template images with known object poses, our method first extracts query embeddings from a query encoder and multiview posed template embeddings from a template encoder. These embeddings serve as conditioning inputs for a diffusion transformer decoder, which is trained to denoise the object pose from random inputs. The model predicts the relative pose between the query and templates, from which the absolute 6D pose of the query object is reconstructed based on the known poses for the templates.
}
    \label{fig:overview}
    \vspace{-0.5cm}
\end{figure*}

The benchmark for object pose estimation (BOP)~\cite{hodan2018bop} has long been dominated by traditional handcrafted feature matching methods based on point pair features (PPF). In recent years, learning-based approaches such as GDR-Net~\cite{wang2021gdrnet}, ZebraPose~\cite{su2022zebrapose}, and SurfEmb~\cite{haugaard2022surfemb} have surpassed traditional methods in performance. However, these methods are instance-specific and require training on each target object. More recently, the community has placed increasing emphasis on \textit{unseen object pose estimation}, which focuses on estimating the pose of novel objects not encountered during training. Below, we describe different pose estimation pipelines used in this setting.

\noindent \textbf{Model-free approaches.} Without 3D model of target objects, Gen6D~\cite{liu2022gen6d}, OnePose~\cite{sun2022onepose}, and OnePose++~\cite{he2022oneposeplusplus} estimate its pose by flipping the structure from motion (SfM) at its head and matching features to align a posed object image to a test view. MFOS~\cite{lee2024mfos} uses posed template images as a model representation and establishes correspondences between the input query image patch and the rendered 3D bounding box of the object associated with each template image. While attractive, this leads to lower pose accuracy caused by this rough bounding box approximation of the object shape.

\noindent \textbf{Template-based approaches.} OSOP~\cite{shugurov2022osop} and OVE6D~\cite{cai2022ove6d} utilize a template object representation for 2D segmentation and coarse to fine matching. MegaPose~\cite{megapose} proposes a generic render-and-compare refinement strategy. GigaPose~\cite{nguyen2024gigapose} performs a template-matching approach in two stages: 1) estimates out-of-plane rotation (2 DoF) by finding discriminative synthetic templates rendered from a CAD model and then 2) establishes correspondences to estimate the four remaining 4 DoF of the object pose. 

\noindent \textbf{Foundation models with CAD model prior.} The idea of foundation models is a recent way to incorporate generic prior knowledge into pose estimation pipelines. Due to the need for an abundance of labeled data, all approaches are trained on synthetic data. Several methods learn generic 3D descriptors such as  Zeropose~\cite{zeropose} and GCPose~\cite{gcpose}. Zeropose predicts poses utilizing the foundation models of ImageBind~\cite{girdhar2023imagebind} and SAM~\cite{kirillov2023segment} together with 3D-3D feature matching. GCPose~\cite{gcpose} uses explicit knowledge of object symmetries. FoundPose~\cite{ornek2024foundpose} combines features from the foundation model DINOv2~\cite{oquab2023dinov2} and bag-of-words retrieval for coarse matching and then uses featuremetric alignment for pose refinement. 
MatchU~\cite{huang2024matchu} and SAM6D~\cite{sam6d} build discriminative descriptors by fusing RGB and depth information using transformers. 
\noindent \textbf{Diffusion in Pose Estimation}
Diffusion models reconstruct a target distribution from noise over multiple time steps, inherently capturing multimodal distributions. They are, by design, capable of capturing multimodal distributions as different noisy initializations can lead to different predictions during inference in the case of a multimodal distribution.
RayDiffusion \cite{zhang2024raydiffusion} denoises camera poses using ray parameterization for multiview estimation, avoiding COLMAP \cite{colmap} in NeRF training but is unsuitable for object pose estimation. Object pose diffusion \cite{hsiao2023se3score} diffuses poses in SE(3) space, excelling in synthetic data but struggling with unseen objects and real datasets. PoseDiffusion \cite{wang2023pd} addresses the SfM problem by diffusing camera poses across multiple images, implicitly performing bundle adjustment. Other methods include DiffusionNOCS \cite{diffusionnocs}, an RGB-D approach that diffuses NOCS maps for pose estimation, and Diff9D \cite{Diff9D}, which estimates 9D pose by diffusing scale, translation, and rotation based on image conditioning.

\section{Method}
\label{sec:method}
\subsection{Method Overview}
In this paper, we represent the 6D object pose using pose maps $\gM$, which encode both orientation and translation. As illustrated in Fig.~\ref{fig:overview}, we adopt a multiview diffusion transformer framework that learns to estimate object pose by denoising noisy pose maps conditioned on an input query object image and a set of reference images with known object poses(termed posed templates).
We extract a query embedding $\gF_Q$ and the multiview template embedding $\gF_{MV}$ using the query and template encoders, respectively. Each encoder consists of an image encoder $\gE_I$ that extracts 2D image features, and a view encoder $\gE_V$ that encodes 6D object pose and/or 2D object location. Specifically, the multiview template features are fused using a Multiview Fuser to form the embedding $\gF_{MV}$. A Diffusion Transformer Decoder is then trained to reconstruct the clean pose maps $\gM_{0}$ from noisy inputs $\gM_{t}$, conditioned on both $\gF_Q$ and $\gF_{MV}$. We train our model with two different template sampling strategies to obtain both coarse and fine pose predictors. For the coarse predictor, template viewpoints are randomly sampled independently of the query pose. For the fine predictor, the same model is trained with templates sampled from a narrower distribution centered around the query pose. This strategy enables coarse-to-fine pose inference during testing without any changes to the network architecture.

\subsection{Object Pose Parameterization}
The 6D object pose is defined by its rotation \( \mR \in SO(3) \) and translation \( \vt \in \mathbb{R}^3 \), representing the transformation from the object's local coordinate frame to the camera coordinate system. While compact pose regression is desirable, it remains challenging for neural networks, especially in generic or cluttered scenes. Recent work~\cite{zhang2024raydiffusion} overparameterizes camera poses using ray directions and ray moments based on Plücker coordinates~\cite{plucker1828analytisch}, which has proven effective for scene-level camera pose estimation. However, this formulation entangles camera intrinsics, rotation, and translation, limiting its effectiveness for object-level pose tasks. Specifically, inaccuracies in the predicted direction map can propagate to the translation component, hindering the centimeter-level precision required in object pose estimation. To overcome this, we propose a novel object-centric representation that maps the 6D object pose into separate 2D rotation and translation maps, enabling more accurate and disentangled learning.

\noindent\textbf{Rotation Parameterization. }
Camera pose estimation or novel view synthesis methods often model camera-centered rays, where rays originate from the camera center and pass through pixel coordinates in the image plane. In contrast, we introduce an object-centered ray representation, where the object center is treated as a \textit{virtual pinhole camera}, emitting rays toward the camera coordinate system. Given the camera intrinsic matrix $\mK \in \mathbb{R}^{3 \times 3}$ and extrinsic parameters—rotation $\mR \in \text{SO}(3)$ and translation $\vt \in \mathbb{R}^3$, a 3D object point $\vx$ is projected onto the image plane as $\vu = \mK [\mR~|~\vt] \vx$.
Instead of relying on this conventional image-based projection, we define a structured representation in which object-centered rays are mapped onto a normalized 2D square grid using a uniform intrinsic matrix, denoted as $\mK = \mK_I$. The set of direction vectors originating from the object center is represented as 
\vspace{-0.2cm}
\begin{equation}
    \mathcal{\mM}_R = \{\vd_1, \ldots, \vd_n\}
\end{equation}
\noindent where each direction vector $\vd_i$ is normalized to unit length. This formulation enables us to map arbitrary rotation matrices $\mR$ onto a unique structured grid on the unit sphere surface. To construct the ray map, we uniformly select $\{\vd_i\}_{i=1}^{n}$ on the projected grid of the sphere surface, ensuring that each vector passes through the center of its corresponding grid cell. Consequently, we obtain a 2D grid map with the shape of $(p \times p \times 3)$ as our rotation representation in the diffusion process. The illustration is given in the supplementary. Given the object-centered ray representation, we recover the rotation matrix $\mR$ by aligning the predicted ray directions with a predefined canonical frame. Let $\mathcal{M}_R = \{\vd_1, \ldots, \vd_n\}$ be the predicted ray set and $\mathcal{M}_R^* = \{\vd_1^*, \ldots, \vd_n^*\}$ the reference rays corresponding to an identity rotation $\mR = \mI$. The optimal rotation matrix $\mR^*$ is obtained by solving:
\vspace{-0.3cm}
\begin{equation}
\mR^* = \argmin_{\mR \in \text{SO}(3)} \sum_{i=1}^{n} \left\| \mR \vd_i^* - \vd_i \right\|^2
\end{equation}
\noindent where $\mR$ is the relative rotation of the object with respect to the canonical frame. This problem can be solved using the Singular Value Decomposition (SVD) differentially, ensuring a valid rotation by enforcing $\mR^T \mR = \mI$. This formulation allows for robust recovery of the object’s orientation and enables the diffusion process on 3D rotations from a structured 2D ray representation.
\begin{figure}[t]
\centering
\includegraphics[width=0.7\columnwidth]{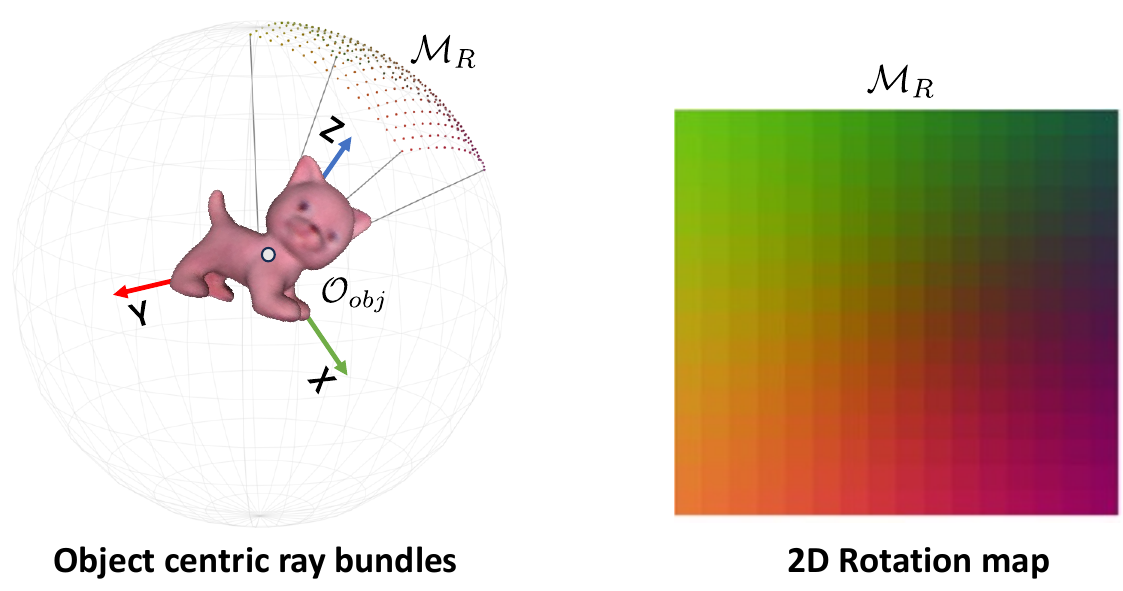}
\vspace{-0.2cm}
    \caption{\small Visual illustration of the object-centric ray representation used for rotation prediction in our diffusion model. The rotation map $\gM_R$ is defined as a bundle of rays originating from the object center $\gO_{obj}$, encoded as a 3-channel 2D map.}
    \label{fig:rotmap}
    \vspace{-0.4cm}
\end{figure}

\noindent\textbf{Translation Parameterization.}
A major challenge in estimating an object's 6D pose from a single RGB image is minimizing translation error, particularly for previously unseen objects and scenes. Earlier work, SSD6D~\cite{kehl2017ssd6d}, estimates translation by locating the object centroid in 2D coordinates and comparing the bounding box scale with a pre-rendered template of the same rotation to determine object distance. However, this approach assumes the object center aligns with the bounding box center, making it sensitive to occlusion. Instance-level regression-based methods~\cite{li2019cdpn,wang2021gdrnet} improve robustness by employing Scale-Invariant Translation Estimation (SITE), which predicts translation by computing the offset between the bounding box center and the object center. More recently, generalizable RGB-based methods~\cite{ornek2024foundpose,nguyen2024gigapose} estimate translation by establishing 2D correspondences between query and template images using a pre-trained feature matcher. While template depth can be rendered, these methods rely solely on one RGB image pair for correspondence extraction. In this paper, we extend SITE to a patch-level dense translation map. Given the object translation $\vt = [\vt_x, \vt_y, \vt_z]$ and the camera intrinsic matrix $\mK$, the projected object centroid $[\vo_x, \vo_y, 1]^T$ in image coordinates is computed as:
\begin{equation}
    [\vo_x, \vo_y, 1]^T = \mK \vt.
\end{equation}
We estimate the offset from each pixel $(u, v)$ in the detected bounding box to the object centroid $(\vo_x, \vo_y)$, forming a dense normalized translation offset map:
\vspace{-0.2cm}
\begin{equation}
    \mathcal{\mM}_T = \left(\frac{u - \vo_x}{w}, \frac{v - \vo_y}{h}, \frac{\vt_z}{r_z} \right),
\end{equation}
\noindent where $w$ and $h$ denote the bounding box width and height, and $r_z$ is the zoom-in ratio of the bounding box. Similar to the rotation map, we uniformly sample the pixels in the bounding box with the same shape of $(p \times p \times 3)$ as the 2D translation map in the diffusion process. The 3D object translation is then recovered by back-projecting the estimated centroid offset using the camera intrinsics:
\begin{equation}
    \vt^* = r_z \cdot \mK^{-1} [w \cdot \Delta \vo_x + \vo_x, h \cdot \Delta \vo_y + \vo_y, \Delta \vo_z]^T.
    \label{equ:translation}
\end{equation}
To this end, we represent object pose as 2D pose maps $\mathcal{\mM} = (\mathcal{\mM}_R, \mathcal{\mM}_T)$. This pose representation decouples rotation and translation as well as the camera intrinsics, enabling the model to predict rotation and translation independently and enabling the use of a diffusion model to denoise the pose on two dense 2D maps. 

\subsection{Multiview Template Conditioned Diffusion}
In our framework, we employ a multiview diffusion model to estimate object pose by conditioning it with the input query image and the posed templates. This network formulates the learning as a denoising process that gradually refines noisy inputs into the proposed structured pose maps.  
\subsubsection{Diffusion Preliminaries}
\paragraph{Diffusion process.} The diffusion process consists of a forward (noising) and a reverse (denoising) process. Given a clean pose representation $\gM_0$ (either the rotation map $\gM_R$ or the translation map $\gM_T$), the forward process adds Gaussian noise over a fixed number of timesteps \( T \). At each timestep \( t \in \{1, \dots, T\} \), the pose map is perturbed as:
\vspace{-0.2cm}
\begin{equation}
    \gM_t = \sqrt{\alpha_t} \gM_0 + \sqrt{1 - \alpha_t} \bm{\epsilon}, \quad \bm{\epsilon} \sim \mathcal{N}(\mathbf{0}, \mathbf{I}),
\end{equation}
\noindent where \( \alpha_t \) is a noise schedule controlling the variance at timestep \( t \).

\noindent\textbf{Denoising process.}
The reverse process aims to recover the clean pose representation by learning to predict and remove the noise. A neural network $\epsilon_\theta(\gM_t, t, \gF_C)$ is trained to estimate the noise $\bm{\epsilon}$ conditioned on an embedding $\gF_c$ that encodes the query and template information. The predicted pose is obtained by iteratively refining $\gM_t$ using the learned noise estimator:
\begin{equation}
\begin{aligned}
    \gM_{t-1} &= \frac{1}{\sqrt{\alpha_t}} 
    \Big( \gM_t - \sqrt{1 - \alpha_t} \epsilon_\theta(\gM_t, t, \gF_c) \Big) \\
    &\quad + \sigma_t \bm{z}, \quad \bm{z} \sim \mathcal{N}(\mathbf{0}, \mathbf{I}).
\end{aligned}
\end{equation}

\noindent where $\sigma_t$ controls the stochasticity of the denoising step. This iterative process gradually refines the noisy pose representation into a structured output.

\subsubsection{Network Architecture}
Our diffusion-based framework for 6D object pose estimation consists of three main blocks: (1) the \textbf{Query Encoder}, which extract the query image features, (2) the \textbf{Template Encoder}, which encodes and fuse the multiple posed template information, and (3) the \textbf{Diffusion Transformer Decoder}, which attends the query and templates and predicts the denoised rotation and translation maps. The overall architecture is illustrated in Figure~\ref{fig:overview}. 

\noindent\textbf{Template Encoder.} The template encoder consists of three components: the \textit{Image Encoder} $\gE_I$, the \textit{View Encoder} $\gE_V$, and the \textit{Multiview Fuser} $\gE_F$. Given $N$ object templates rendered from different viewpoints, we first employ a frozen DINOv2~\cite{oquab2023dinov2} backbone to extract the image feature maps $\gF_T$. Inspired by prior works~\cite{nerf, PerceiverIO, TowardsZero, ZeroshotNovelView}, which have demonstrated the effectiveness of Fourier encoding for camera rays in multiview scene understanding and reconstruction, we extend this idea to embed our structured pose information.

\noindent\textit{\textbf{View Encoder.}} In our framework, the \textit{View Encoder} uses three Fourier encoders to process the structured rotation map, translation map, and the normalized bounding box coordinates in the 2D image. This view embedding $\mathcal{\mF}_v$ explicitly locates the objects by incorporating both their viewpoint in 3D space and their scale in 2D image coordinates, enforcing the network to learn implicit relationships across different views. The Fourier feature embedding for a scalar input $x$ is computed as $\gamma(x) = \left(x, \sin(2\pi B x), \cos(2\pi B x) \right)$, where $B$ is a fixed frequency band that controls the resolution of the encoded features. Given the bounds $K_r$, $K_t$, and $K_c$ for rotation, translation, and object 2D coordinate maps, the total dimensionality of the view embedding is $D_V = (2(K_r + K_t + K_c)+1)d$, where $d$ is the number of frequency bands used for each Fourier encoding. This formulation ensures that each structured feature is transformed into a high-dimensional space, improving the model’s ability to capture fine-grained variations in object pose.

\noindent\textit{\textbf{Multiview Fuser.}} The view embeddings $\gF_{v}$ and the image embeddings $\gF_{I}$ are concatenated and conditioned with view-level and patch-level positional encodings. This combined representation is then passed to the \textit{Multiview Fuser}, which consists of $\gN_F$ self-attention layers to extract the fused multiview template embedding $\gF_{MV}$. Following~\cite{zhang2024raydiffusion}, we use DiT~\cite{dit} blocks for cross-view information exchange and ensure that the network effectively aggregates object appearance and viewpoint-dependent geometric cues across all template images. The multiview embedding encodes the 2D and 3D priors of the object implicitly, which will be used as conditioning information to help reason the pose of the query image. 

\noindent\textbf{Query Encoder.} The Query Encoder shares the same image and view encoders as the Template Encoder but processes only a single-view input. Since the pose of the query is not known, only the object's 2D location in image coordinates is conditioned, instead of the full pose. Patch-level positional encoding is also incorporated to capture the spatial position of patches in the query image, facilitating implicit fusion with template patches. The resulting query embedding is denoted as $\gF_{Q}$.

\noindent\textbf{Diffusion Transformer Decoder.}
The fused multiview template embedding $\gF_{MV}$ and the query embedding $\gF_{Q}$ serve as conditioning information in our diffusion transformer decoder, which denoises the 2D rotation and translation maps over a series of timesteps. At each step $t$, the noisy pose maps $\gM_t$ are processed alongside their conditional embeddings through a sequence of $\gN_D$ transformer-based diffusion blocks based on DiT~\cite{dit}. Unlike DiT, which primarily uses only self-attention layers, we follow~\cite{li2024hunyuan} to also incorporate cross-attention layers, allowing the query embedding to attend to the fused template embedding, enabling the feature exchange between different latent spaces that are constructed from a single query and multiple posed template views respectively. During training, for each time step $t$, the decoder learns to predict the noise component $\epsilon_\theta(\gM_t, t, \gF_C)$, where $\gF_C = (\gF_{Q}|\gF_{MV})$. During inference, the pose maps are randomly initialized and iteratively denoised to obtain the final 2D rotation and translation maps. To enhance generalization, instead of directly predicting the absolute pose of the query, we estimate the relative pose between the query and templates. Specifically, the rotation maps represent the relative rotation from the template to the query, while the translation maps encode a relative depth scale between them: 
\vspace{-0.3cm}
\begin{equation}
    \vt^{rel}_z = \frac{\vt^{Q}_{z} r^{T}_{z}}{\vt^{T}_{z} r^{Q}_{z}}
\end{equation} \vspace{-0.1cm}
\noindent where $\vt^{rel}_z$ is the predicted relative depth scale, $\vt^{Q}_{z}$ and $\vt^{T}_{z}$ are the ground-truth depths of the object center for the query and templates respectively, and $r^{Q}_{z}$ and $r^{T}_{z}$ are the zoom-in scales of the query and template. Finally, the query pose $\mH_Q = [\mR_Q|\vt_Q]$ is recovered from the predicted relative pose $\mH_{rel}$ and the ground-truth pose $\mH_{gt}$ of the templates:
\vspace{-0.3cm}
\begin{equation}
    \mH_Q = \mH_{rel} \mH_{gt}.
\end{equation}

Our multiview DiT-based Diffusion Transformer Decoder estimates the relative pose between a query and each template by conditioning on implicit geometric priors encoded in the templates and the query features. Each template view produces an independent pose hypothesis, supervised by its specific relative pose to the query. This design enables efficient, batch-wise probabilistic sampling from randomly selected templates, yielding diverse pose hypotheses in parallel.

\subsection{Loss Functions} 
In our diffusion model, instead of training the network to estimate the noise, we follow prior work~\cite{zhang2024raydiffusion} and train the denoising network $\epsilon_\theta(\gM_t, t, \gF_c)$ to learn the reverse diffusion process by predicting the original clean pose map $\gM_0$ conditioned on the noisy input $\gM_t$ at timestep $t$. The loss function is defined as:
\begin{equation}
    \gL_{\text{diff}} = \mathbb{E}_{t,\epsilon}\left[\|\gM_0 - \epsilon_\theta(\gM_t, t, \gF_c)\|_2^2\right],
\end{equation}
where $t$ is uniformly sampled from $[1, T]$ during training, and $\gM_t$ is the noisy version of the original pose map $\gM_0$ corrupted with Gaussian noise at timestep $t$. This training strategy naturally integrates task-specific constraints on the original rotation and translation maps.

\noindent \textbf{Rotation losses.}  
For the rotation map, we apply a pixel-level reconstruction loss to the target map $\gM_R^*$:
\vspace{-0.3cm}
\begin{equation}
    \gL^R_{\text{recon}} = \frac{1}{p^2}\|\gM_R - \gM_R^*\|^2_2,
\end{equation}
where $p$ denotes the spatial resolution of the rotation ray maps. Since each element in the rotation map represents a directional vector, we also employ a cosine similarity loss $\gL^R_{\text{cos}}$ to supervise ray directions. Given that the rotation map consists of structured ray bundles, we introduce an angle-consistency loss to enforce geometric coherence across adjacent rays in the predicted ray map. Given the predicted ray directions $\{\vd_{i}\}_{i=1}^{n}$ and the canonical ray set $\{\vd_{i}^*\}_{i=1}^{n}$, we ensure that the predicted rays maintain consistent relative angles that reflect the intrinsic ray map geometry. For each pair of adjacent rays indexed by $(i,j)$, the relative angle is computed as $\alpha_{ij} = \arccos(\vd_i^\top \vd_j)$. Similarly, we precompute the reference angles $\alpha_{ij}^*$ from the canonical rays corresponding to an identity rotation. The ray-consistency loss is then defined as:
\begin{equation}
\gL^R_{\text{reg}} = \frac{1}{|\gN_r|} \sum_{(i,j) \in \gN_r} \left(\alpha_{ij} - \alpha_{ij}^*\right)^2,
\end{equation}
\noindent where $\gN_r$ is the set of adjacent ray index pairs, and $|\gN_r|$ is its cardinality. This loss term encourages the network to respect intrinsic geometric constraints imposed by the projection, ensuring stable rotation estimation. The overall rotation loss is then formulated as a weighted combination of the reconstruction loss, cosine similarity loss, and ray-consistency loss:
\begin{equation}
\gL^R = \lambda_{\text{recon}} \gL^R_{\text{recon}} + \lambda_{\text{cos}} \gL^R_{\text{cos}} + \lambda_{\text{reg}} \gL^R_{\text{reg}},
\end{equation}
\noindent where $\lambda_{\text{recon}}, \lambda_{\text{cos}},$ and $\lambda_{\text{reg}}$ are hyperparameters that balance the contributions of each term. This formulation ensures that the predicted rotation map is both accurate at the pixel level and geometrically consistent with the camera's intrinsic structure.

\noindent \textbf{Translation losses.}  
For the translation map, similar to the rotation ray reconstruction, we apply a pixel-level L2 loss to ensure accurate reconstruction of the dense translation offset maps:
\vspace{-0.3cm}
\begin{equation}
    \gL^T_{\text{recon}} = \frac{1}{p^2}\|\gM_T - \gM_T^*\|^2_2.
\end{equation}
Additionally, to explicitly supervise the final 3D translation prediction, we compute the object translation $\vt$ from the predicted translation map using the decoding formulation described in Eq.~\ref{equ:translation}. We then impose an L1 loss on the predicted translation components along each axis:
\begin{equation}
\gL^T_{\text{xyz}} = \lambda_x|\vt_x - \vt_{x}^*| + \lambda_y|\vt_y - \vt_y^*| + \lambda_z|\vt_z - \vt_z^*|,
\end{equation}
\noindent where $\lambda_x, \lambda_y, \lambda_z$ are weighting the supervision strength. The overall translation loss is defined as the weighted sum of these two terms:
\begin{equation}
    \gL^T = \lambda_{\text{recon}}\gL^T_{\text{recon}} + \lambda_{t}\gL_{\text{xyz}},
\end{equation}
where $\lambda_{\text{recon}}$ and $\lambda_{t}$ are the hyperparameters to balance dense map-level supervision and explicit object-level translation regression.

\noindent \textbf{Overall loss function.}  
The final training loss is defined as a weighted sum of the rotation and translation losses:
\begin{equation}
    \gL = \lambda_{\text{rot}}\gL^R + \lambda_{\text{trans}}\gL^T,
\end{equation}
\noindent where $\lambda_{\text{rot}}$ and $\lambda_{\text{trans}}$ are hyperparameters controlling the relative importance of rotation and translation losses.

\subsection{Coarse to Fine Predictor Training Strategy.}
We train our network with different template distributions to obtain both a coarse and fine predictor. For the coarse predictor, we use eight templates with randomly sampled poses from pre-processed scene-cropped images, ensuring diverse viewpoint coverage. The fine predictor employs online template sampling, augmenting templates for $\pm30^\circ$, $\pm5$cm based on the query pose to enforce closer template-query alignment. Both predictors share the same network architecture but differ in template distributions during training. During inference, the coarse predictor provides an initial pose estimate, which is then refined by the fine predictor using a more localized template distribution.

%-------------------------------------------------------------------------

\section{Experiments}
\label{sec:experiment}
\begin{table*}[t]
    \centering
    \resizebox{0.8\linewidth}{!}{%
    \begin{tabular}{lcccccccc}
        \toprule
        \textbf{Method} & \textbf{Refinement} & \textbf{Multi-hypo} & \textbf{LM-O~\cite{brachmann2014learning}} & \textbf{T-LESS~\cite{hodan2017tless}} & \textbf{TUD-L~\cite{hodan2018bop}} & \textbf{IC-BIN~\cite{doumanoglou2016recovering}} & \textbf{YCB-V~\cite{xiang2018posecnn}} & \textbf{Average}\\
        \midrule
        OSOP~\cite{shugurov2022osop}      & \cross & \cross & 31.2 & --   & --   & --   & 33.2 & 32.2 \\
        ZS6D~\cite{ausserlechner2024zs6d}      & \cross & \cross & 29.8 & 21.0 & --   & --   & 32.4 & 27.7 \\
        MegaPose~\cite{megapose}  & \cross & \cross & 22.9 & 17.7 & 25.8 & 15.2 & 28.1 & 21.9 \\
        GenFlow~\cite{moon2024genflow}   & \cross & \cross & 25.0 & 21.5 & 30.0 & 16.8 & 27.7 & 24.2 \\
        GigaPose~\cite{nguyen2024gigapose}  & \cross & \cross & 29.9 & 27.3 & 30.2 & 23.1 & 29.0 & 27.9 \\
        FoundPose~\cite{foundationpose} & \cross & \cross & 39.6 & 33.8 & 46.7 & \textbf{23.9} & 45.2 & 37.8 \\
        Ours      & \cross & \cross & \textbf{42.1}  & \textbf{36.9} & \textbf{48.3} & 21.8 & \textbf{46.2} & \textbf{39.1} \\
        \midrule
        \midrule
        MegaPose~\cite{megapose}  & \checkmark & \cross & 49.9 & 47.7 & 65.3 & 36.7 & 60.1 & 51.9 \\
        GigaPose~\cite{nguyen2024gigapose}  & \checkmark & \cross & 55.6 & \textbf{54.6}  & 57.8 & \textbf{44.3}  & 63.4 & 55.1 \\
        FoundPose~\cite{foundationpose} & \checkmark & \cross & 55.7 & 51.0 & 63.3 & 43.3 & 66.1 & 55.9 \\
        Ours      & \checkmark & \cross & \textbf{56.2}  & 53.8  & \textbf{66.5} & 41.6 & 62.8 & \textbf{56.2} \\
        \midrule
        \midrule
        GenFlow~\cite{moon2024genflow}   &\checkmark  & \checkmark   & 56.3 & 52.3 & 68.4 & 45.3 & 63.3 & 57.1 \\
        MegaPose~\cite{megapose}  &\checkmark  & \checkmark   & 56.0 & 50.7 & 68.4 & 41.4 & 62.1 & 55.7 \\
        GigaPose~\cite{nguyen2024gigapose}  & \checkmark & \checkmark   & 59.9 & 57.0 & 64.5 & 46.7 & 66.3 & 58.9 \\
        FoundPose~\cite{foundationpose} & \checkmark & \checkmark   & 61.0 & 57.0 & 69.4 & \textbf{47.9} & \textbf{69.0} & 60.9 \\
        Ours      & \checkmark & \checkmark   & \textbf{62.2} & \textbf{59.1} & \textbf{70.2}  & 45.5  & 68.9 & \textbf{61.2} \\
        \bottomrule
    \end{tabular}
    }\vspace{-0.2cm}
    \caption{\small We compare our method against RGB-only baselines by reporting the Average Recall (AR) scores on five BOP core datasets.}
    \label{tab:results}
    \vspace{-0.4cm}
\end{table*}

\subsection{Experimental Setup}
\noindent\textbf{Evaluation Metrics.} We adopt the metric Average Recall (AR) proposed by the Benchmark of Pose Estimation (BOP)~\cite{sundermeyer2023bop}. The AR score is calculated with 3 pose-error functions: Visible Surface Discrepancy (VSD), Maximum Symmetry-Aware Surface Distance (MSSD), and Maximum Symmetry-Aware Projection Distance (MSPD). A pose is considered correct if the pose errors are within a predefined error threshold. The mean recall on the each error functions is computed over multiple error thresholds. The overall accuracy of a method is given by the Average Recall $AR = (AR_{\text{VSD}} + AR_{\text{MSSD}} + AR_{\text{MSPD}})/{3}$.

\noindent\textbf{Training and evaluation datasets.}
We train our model on realistic synthetic datasets generated by Megapose~\cite{megapose}, comprising approximately 2 million images rendered with BlenderProc~\cite{denninger2019blenderproc} using objects from Google Scanned Objects~\cite{gso} and ShapeNet~\cite{chang2015shapenet}. For novel object pose estimation, we evaluate our method on five benchmark datasets: LM-O~\cite{brachmann2014learning}, T-LESS~\cite{hodan2017tless}, YCB-V~\cite{xiang2018posecnn}, TUD-L~\cite{hodan2018bop}, and IC-BIN~\cite{doumanoglou2016recovering}. Our evaluation is structured as follows: in Section~\ref{sec:baselines}, we compare our method with baselines on novel object pose estimation; in Section~\ref{sec:ablation}, we conduct ablation studies where we analyze design components, where we train and evaluate our method on LM-O dataset.

\subsection{Compare to Baselines}
\label{sec:baselines}
We evaluate our method on five benchmark datasets: LM-O, T-LESS, TUD-L, IC-BIN, and YCB-V, which are unseen during training, and compare it with recent state-of-the-art methods that use only RGB images as input. All the methods use the same detection and segmentation results generated from CNOS~\cite{nguyen2023cnos} by default, except for OSOP~\cite{osop}. As shown in Table~\ref{tab:results}, we analyze different setups, considering whether refinement and multi-hypothesis predictions are used. Our method achieves the highest average AR across all settings. In the single-prediction setting without refinement, it improves over the previous best method by 3.4\% on average, with notable gains of 6.3\% on LM-O and 9.2\% on T-LESS. With refinement, our method continues to outperform the baselines, particularly excelling on TUD-L and LM-O. In the multi-hypothesis setting, it achieves the best performance on most datasets, especially with T-LESS dataset being improved by 3.7\%. These results highlight the effectiveness of our approach in enhancing pose estimation by leveraging robust pose representations and a diffusion-based pipeline while ensuring strong generalization across diverse datasets.

\begin{table}[t]
    \centering
    \resizebox{\columnwidth}{!}{%
    \begin{tabular}{l c c c c c}
        \toprule
        \textbf{Method} & \textbf{GT Temp. Pose}  & \textbf{Multiview} & \textbf{Relative Pose} & \textbf{Fine Predictor} & \textbf{AR} \\
        \midrule
    (1) \textbf{\textit{Fine+MegaPose}} & \checkmark  & \checkmark & \checkmark & \checkmark & \textbf{56.2} \\
    (2) \textit{Fine}    & \checkmark &  \checkmark & \checkmark & \checkmark & 42.1 \\
    (3) \textit{Coarse}  & \checkmark & \checkmark & \checkmark & \cross & 39.6 \\
    \midrule
    (4) \textit{Absolute pose} &  \checkmark & \checkmark & \cross  &\cross & 35.4 \\
    (5) \textit{Single view} &  \checkmark & \cross & \cross & \cross & 32.5 \\
    (6) \textit{w/o GT Template Pose} & \cross & \checkmark & \checkmark & \cross & 27.8 \\
        \bottomrule
    \end{tabular}%
    }\vspace{-0.2cm}
    \caption{ Ablation study for the key components of our method.} \vspace{-0.4cm}
    \label{tab:ablation}
\end{table}

\subsection{Ablation Study}
\label{sec:ablation}
We conduct an ablation study on four key components of our approach: template ground-truth (GT) view embedding, the multiview setup, the fine-level predictor, and relative pose prediction. Each component is either removed or replaced with an alternative setup, and the results are summarized in Table~\ref{tab:ablation}. 

\noindent\textbf{Fine predictor.}  
In the refinement stage, we apply both our fine predictor and MegaPose refinement. As shown in (1) – (3) of Table~\ref{tab:ablation}, our fine predictor improves performance by 6.3\% compared to the coarse prediction. Notably, the fine predictor does not modify the network itself but instead utilizes a different template sampling strategy. Further improvements are achieved when incorporating an external refiner during the refinement stage.

\noindent\textbf{Relative pose prediction.}  
To enhance generalization across different scenes, camera intrinsics, and viewing conditions, we predict the relative pose between posed templates and the query, using the ground-truth template pose to infer the absolute query pose. In this ablation, we modify the network to directly predict the absolute poses of all template and query frames using a sequence of DiT blocks. As shown in (4) of Table~\ref{tab:ablation}, this modification results in a performance drop of up to 10\%, highlighting the effectiveness of relative pose prediction.

\noindent\textbf{Multi-view prediction.}  
We modify our pose map prediction head to perform single-view prediction, meaning the query pose is estimated directly without leveraging multiple templates. As shown in (5) of Table~\ref{tab:ablation}, this change leads to an approximately 18\% performance drop, confirming that multi-view prediction enables the model to learn stronger implicit correspondences between the query and templates.

\noindent\textbf{Ground-truth template pose.}  
In our setup, we explicitly input template ground-truth pose maps as conditional information to help the network learn inherent correlations with the input query image. To evaluate its impact, we remove the ground-truth pose map from the input while retaining only the 2D position information, which is essential for predicting the relative distance from the camera. The ground-truth template pose is only used during inference to recover the absolute query pose. As shown in (6) of Table~\ref{tab:ablation}, removing the template ground-truth pose map results in a significant performance drop compared to the single-view setting. This finding underscores the importance of leveraging template pose priors in multi-view prediction, and also indicates the effectiveness of the view encoders. 
\vspace{-0.3cm}

\begin{table}[t]
    \centering
    \resizebox{0.6\columnwidth}{!}{%
    \begin{tabular}{l c c}
        \toprule
        \textbf{Method} & \textbf{Template Distribution} & \textbf{AR} \\
        \hline
        Coarse & Random & 65.58 \\
        Coarse & fixed  & 60.28 \\
        \midrule
        Fine & $\pm90^\circ$, $\pm10$cm & 65.81 \\
        \textbf{Ours-Fine} & $\pm30^\circ$, $\pm5$cm & \textbf{67.29} \\
        Fine & $\pm15^\circ$, $\pm3$cm & 61.04 \\
        \bottomrule
    \end{tabular}
    } \vspace{-0.2cm}
    \caption{\small Comparison of different template selection strategies for the fine predictor on LM-O dataset. The bold is our default setup.}\vspace{-0.6cm}
    \label{tab:template_distribution}
\end{table}

\subsection{Effects of pose distribution of templates.} We evaluate our method's sensitivity to template distribution by training with different template sampling strategies on the LM-O dataset, as shown in Table~\ref{tab:template_distribution}. For the coarse predictor, using randomly sampled templates yields better performance than the fixed-template setting. For the fine predictor, we examine different template distributions with varying pose and translation constraints. The best performance is achieved with a template distribution constrained to $\pm30^\circ$ in rotation and $\pm5$cm in translation, which aligns with the mean error of the coarse predictor. A wider distribution ($\pm90^\circ$, $\pm10$cm) performs similarly to the randomly sampled distribution used in the coarse predictor, while overly narrow constraints ($\pm15^\circ$, $\pm3$cm) lead to a slight performance drop. These results underscore the importance of selecting appropriate template distributions for both coarse and fine predictors to balance generalization and fine-level accuracy.

\vspace{-0.2cm}\section{Conclusion}
In this paper, we introduced a structured representation for object pose that enables effective deployment of diffusion models for object 6D pose estimation. Instead of pairwise matching, we propose aligning object-centered rays across multiple posed templates. Our multiview diffusion model is conditioned on embeddings extracted from both the query and multiple posed template images using dedicated encoders. A coarse-to-fine strategy refines pose accuracy without architectural changes, allowing probabilistic reasoning over multiview inputs without explicit 3D reconstruction. While achieving competitive performance, the approach relies on posed templates and accurate detections. Future work may focus on relaxing these constraints for broader generalization.

{
    \small
    \bibliographystyle{ieeenat_fullname}
    \bibliography{main}
}

\end{document}